\documentclass[pdflatex,sn-mathphys-num]{sn-jnl}

\usepackage{graphicx}
\usepackage{multirow}
\usepackage{amsmath,amssymb,amsfonts}
\usepackage{amsthm}
\usepackage{mathrsfs}
\usepackage[title]{appendix}
\usepackage{xcolor}
\usepackage{colortbl}
\usepackage{textcomp}
\usepackage{manyfoot}
\usepackage{booktabs}
\usepackage{algorithm}
\usepackage{algorithmicx}
\usepackage{algpseudocode}
\usepackage{listings}

\newcommand{\cmark}{\checkmark}
\newcommand{\xmark}{$\times$}
\newcommand{\pmark}{$\circ$}
\definecolor{hmhigh}{HTML}{DDECCB}
\definecolor{hmmid}{HTML}{EEF4E7}
\definecolor{hmlow}{HTML}{F7F0D8}
\definecolor{hmwarn}{HTML}{F4DAD4}
\newcommand{\hmhighcell}[1]{\cellcolor{hmhigh}\textbf{#1}}
\newcommand{\hmmidcell}[1]{\cellcolor{hmmid}#1}
\newcommand{\hmlowcell}[1]{\cellcolor{hmlow}#1}
\newcommand{\hmwarncell}[1]{\cellcolor{hmwarn}#1}

\theoremstyle{thmstyleone}

\theoremstyle{thmstyletwo}

\theoremstyle{thmstylethree}

\makeatletter
\renewcommand{\@@address}[2][]{%
    \stepcounter{affn}%
    \xdef\@currentlabel{\theaffn}%
    \jmkLabel{\theaffn}%
    \ifnum\addcount=1\relax%
        \g@addto@macro\auaddress{\textsuperscript{#1}#2}%
    \else%
        \g@addto@macro\auaddress{;\ \textsuperscript{#1}#2}%
    \fi%
}
\renewcommand{\presentaddress}[1]{\gdef\@presentaddresstext{$^{\ddagger}$#1}\global\presentaddresstrue}
\renewcommand{\@maketitle}{\newpage\null%
    \if@remarkboxon\vbox to 0pt{\vspace*{-78pt}\hspace*{-18pt}\FMremark}\else\vskip21pt\fi%
    \hsize\textwidth\parindent0pt%
    {\hbox to \textwidth{{\Artcatfont\ArtType\hfill}\par}}
    \ifx\@title\empty\else%
        \removelastskip\vskip20pt\nointerlineskip%
        {\Titlefont\@title\par}
    \fi%
    \ifx\@subtitle\empty\else%
        \vskip9pt%
        {{\SubTitlefont\@subtitle\par}}
    \fi%
    \ifnum\aucount>0
        \global\punctcount\aucount%
        \vskip20pt%
        \artauthors\par%
        {\vskip8pt{\reset@font\fontsize{9bp}{11bp}\selectfont\titraggedcenter\auaddress\par}%
        \removelastskip\vskip12pt%
        {\reset@font\fontsize{9bp}{11bp}\selectfont\titraggedcenter%
        \ifnum\emailcnt>0\relax%
           \ifx\corrauthemail\@empty\else{\ifnum\aucount>1*\fi}Corresponding authors: \corrauthemail\fi%
           \ifx\authemail\@empty\else Contributing authors: \authemail\fi%
        \fi%
        \ifequalcont{\quad$^{\dagger}$\@equalconttext}\fi%
        \ifpresentaddress{\quad\@presentaddresstext}\fi%
        \par}%
        \removelastskip\vskip14pt%
        }
     \fi%
     {\printabstract\par}%
     {\printkeywords\par}%
     \ifx\@pacs\empty\else%
       \loop\ifnum\PacsCount>0%
          \csname\romannumeral\PacsTmpCnt StorePacsTxt\endcsname\par%
          \StepDownCounter{\PacsCount}%
          \StepUpCounter{\PacsTmpCnt}%
       \repeat%
    \fi%
    \ifx\@motto\empty\else\@motto\fi%
}
\makeatother

\begin{document}

\title[Adaptive Questions and World-Model Probes]{Learning Explicit Behavioral Models with Adaptive Questions and World-Model Probes}

\author[1,3]{\fnm{Hikaru} \sur{Shindo}}\equalcont{These authors contributed equally to this work.}
\author*[1,2,3]{\fnm{Yu} \sur{Deng}}\email{yu.deng@tu-darmstadt.de}\equalcont{These authors contributed equally to this work.}
\author*[1,2,3]{\fnm{Teng} \sur{Cao}}\email{teng.cao@tu-darmstadt.de}\equalcont{These authors contributed equally to this work.}
\author[]{\fnm{Quentin} \sur{Delfosse}\textsuperscript{1,3,\ensuremath{\ddagger}}}
\author[1,2,3]{\fnm{Christopher} \sur{Tauchmann}}
\author[1,3]{\fnm{Jannis} \sur{Bl{\"u}ml}}
\author[1,3]{\fnm{Gopika} \sur{Sudhakaran}}
\author[1,3,2,4,5]{\fnm{Kristian} \sur{Kersting}}

\affil[1]{\orgdiv{Artificial Intelligence and Machine Learning Lab}, \orgname{Technical University of Darmstadt}, \orgaddress{\country{Germany}}}
\affil[2]{\orgname{Hessian Center for Artificial Intelligence (hessian.AI)}, \orgaddress{\country{Germany}}}
\affil[3]{\orgdiv{Department of Computer Science}, \orgname{Technical University of Darmstadt}, \orgaddress{\country{Germany}}}
\affil[4]{\orgname{German Research Center for Artificial Intelligence (DFKI)}, \orgaddress{\country{Germany}}}
\affil[5]{\orgdiv{Centre for Cognitive Science}, \orgname{Technical University of Darmstadt}, \orgaddress{\country{Germany}}}
\presentaddress{Quentin Delfosse is now at Intrinsic AI, Google.}
\makeatletter
\gdef\corrauthemail{\textcolor{blue}{teng.cao@tu-darmstadt.de};\ \textcolor{blue}{yu.deng@tu-darmstadt.de};\ }
\makeatother

\abstract{Interactive agents trained only against task return can achieve high scores while failing to represent the mechanisms that make their actions succeed. This makes brittle behavior difficult to diagnose and limits adaptation when environment dynamics change. Existing LLM reflection and policy-code repair can revise behavior from failed trajectories, but questions and world-understanding tests are usually used only after training. We introduce an Explicit Symbolic Behavioral Model (ESBM), a trainable behavioral model that couples task performance with evidence-grounded question answering and executable mechanism prediction. An ESBM represents behavior through typed predicates, weighted rules, bounded options and mechanism memory; the mechanism layer predicts symbolic events, object changes, rewards and terminal consequences under action interventions. After each rollout, adaptive questions and active world-model probes convert score failures, QA errors and transition-prediction errors into constraints for local ESBM edits. Candidate models are selected by a multi-criterion rule that jointly evaluates task score, answerability and active world-model consistency. Under the tested Atari-style protocols, ESBM learns high-scoring policies while producing explicit answers and executable mechanism predictions, indicating that adaptive questions can serve as both training pressure and reusable benchmarks for mechanistic policy learning in this setting.}


\maketitle

\section{Introduction}\label{sec:introduction}

Interactive agents are usually evaluated by what they achieve, not by whether they represent why their actions work. In games and other sequential environments, this distinction is not cosmetic. A policy may obtain high reward by exploiting a narrow timing pattern, repeatedly collecting a local reward, or following an environment-specific shortcut, while lacking a usable model of the environment mechanisms that make the behavior succeed. Such high-score but low-understanding behavior is difficult to detect from scalar reward alone and is a common source of brittle policies.

\begin{figure}[t]
\centering
\includegraphics[width=\textwidth]{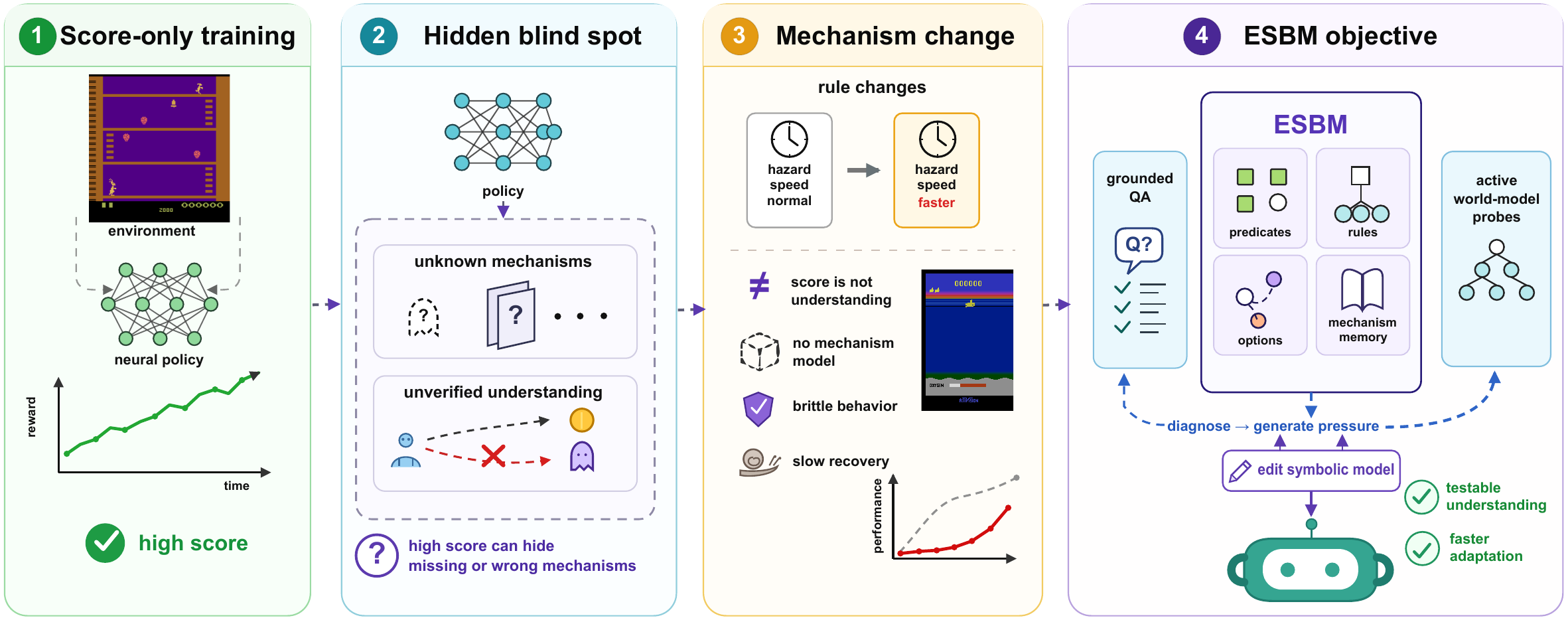}
\caption{\textbf{Problem definition and motivation.}
Task return alone can make a policy appear successful even when the mechanisms behind its behavior remain implicit or wrong. Under mechanism changes, score-only behavior may reveal brittle adaptation and slow recovery. ESBM addresses this gap by making predicates, rules, options and mechanism memory explicit objects for grounded questions, active world-model probes and local symbolic edits.}
\label{fig:problem-definition}
\end{figure}

Figure~\ref{fig:problem-definition} illustrates this problem definition. A score-only policy can appear successful while its mechanisms remain unknown or unverified; when the environment mechanism changes, the same policy may become brittle and recover slowly. This motivates optimizing an explicit behavioral model whose mechanisms can be queried, checked and locally edited.

Recent LLM-based agents provide a natural way to revise policies from experience. Reflection-based agents can summarize failures and use verbal feedback to improve behavior \cite{reflexion,selfrefine}. Program-editing and embodied-control systems can generate executable skills or policy code from language and environment feedback \cite{voyager,codeaspolicies}. Related work also uses LLMs to generate reward functions or dense training signals \cite{eureka,text2reward}. More recent self-evolving agents co-train a curriculum generator and an executor from model-generated tasks \cite{agent0}. These systems show that LLMs can inspect failed trajectories, generate challenging tasks and revise executable artifacts, but the revision loop often still relies on task success or latent model improvement as the primary judge of whether the change is useful.

In parallel, work on world-understanding evaluation and dynamic benchmarks argues that aggregate task success is an incomplete proxy for grounded knowledge. Question-answering probes can test whether an agent knows object identities, hazards, reward mechanisms, action consequences, and counterfactual dynamics. Dynamic and adversarial benchmarks further show that fixed test sets can saturate and fail to expose model-specific blind spots \cite{dynabench,adversarialqa,anli,adversarialvqa}. However, these benchmarks are commonly used after training. They rarely become part of the mechanism that selects, repairs, or rejects a policy.

This paper asks whether understanding can be optimized as part of the policy-learning loop. The central difficulty is that ``understanding'' must be testable rather than rhetorical. We therefore treat it as an explicit and executable object: a model state that can answer grounded questions, predict symbolic consequences and influence action selection. The agent is represented as an Explicit Symbolic Behavioral Model (ESBM), consisting of state-abstraction predicates, weighted action rules, bounded options and mechanism memory. The mechanism memory is not only a natural-language explanation store; it is an abstract world model that predicts how symbolic state, reward, lives and objects change under actions.

We propose adaptive questions and world-model probes as a training mechanism for ESBMs. After each rollout, a challenger uses rollout failures, recent symbolic edits, QA mistakes and transition-prediction errors to generate two forms of pressure. The first is grounded QA, which asks about entities, hazards, rewards, routes, action consequences and counterfactual changes. The second is active world-model probing, which branches the real environment and the ESBM's abstract world model from matched checkpoints under the same observed, counterfactual or mixed action sequence. The abstract model must then predict future events, score changes, life changes, player motion and object changes.

The policy representation is deliberately symbolic. Predicates, rules, options, and mechanisms make updates local, executable, and auditable. A candidate change can be traced to a predicate threshold, a rule condition, an option contract, or a mechanism transition. This substrate creates an interpretable chain from failure trace, to generated probe, to understanding failure, to ESBM edit, to changed behavior. It also makes high-score but low-understanding behavior observable: a candidate can improve task return while still being exposed by mechanism QA or active transition checks, and the accepted model can later be stress-tested by continuing optimization after controlled mechanism changes.

We evaluate the approach on three primary Atari-style JAXAtari environments: \texttt{kangaroo}, \texttt{seaquest} and \texttt{kingkong}. These games require route planning, resource management, rescue behavior, timing and hazard avoidance, so they test both task performance and mechanism understanding. The experiments ask four questions: whether ESBM achieves competitive task scores, whether the learned model answers evidence-linked game-understanding questions, whether its mechanisms predict executable symbolic dynamics, and whether the model recovers after mechanism changes. Additional game scores are reported in the appendix for broader score context beyond the main comparison set.

Our contributions are:
\begin{itemize}
    \item We define the ESBM as an explicit optimization object that combines state abstraction, symbolic action selection, bounded options, and executable mechanism memory.
    \item We introduce policy-adaptive, evidence-grounded questions and active world-model probes that convert the current model's failures into reusable understanding constraints.
    \item We formulate a multi-criterion acceptance rule that selects local ESBM edits using task performance, QA accuracy and active interventional world-model consistency.
\end{itemize}

\section{Background}\label{sec:background}

Classical symbolic policies use predicates to map observations into semantic facts and weighted rules to map those facts into actions. This representation is interpretable, but a flat predicate-rule policy is not enough for the setting studied here: some behavior requires memory or bounded multi-step procedures, and a correct action rule may still lack an explicit account of the mechanism that makes it correct. We therefore treat the policy as one component of a larger explicit behavioral model.

LLM agents can revise behavior from rollouts, execution traces and failed attempts. Prior work has explored reflection, self-feedback, executable skill repair, policy-code generation and reward-code generation \cite{reflexion,selfrefine,voyager,codeaspolicies,eureka,text2reward}. We use this ability as an editing backend, but select edits with score, QA and executable world-model validation rather than with task reward alone.

Score measures task outcome, whereas many failures are mechanistic: the agent may misidentify a hazard, misunderstand a reward source, choose an action with the wrong consequence or follow a route that only works under the original dynamics. QA probes these mechanisms in language, and world-model checks test whether mechanism memory predicts symbolic consequences under state-action interventions.

\section{Problem Formulation}\label{sec:problem}

At iteration $t$, the agent is an ESBM
\begin{equation}
    M_t = (\Sigma, \Phi_t, \Pi_t, K_t, O_t),
\end{equation}
where $\Sigma$ is the symbolic vocabulary, $\Phi_t$ maps observations and memory to facts, $\Pi_t$ maps facts to actions or option calls, $K_t$ is executable mechanism memory, and $O_t$ is a bounded option library. Given observation $o_t$ and memory $m_t$, execution is
\begin{equation}
    z_t=\Phi_t(o_t,m_t,K_t), \qquad
    u_t=\Pi_t(z_t), \qquad a_t,m_{t+1}=\operatorname{Execute}(u_t,o_t,m_t,O_t).
\end{equation}

The loop summarizes the information available for revision as
\begin{equation}
    S_t = \{M_t, E_t, D_t, Q_t, H_t, U_t\},
\end{equation}
where $E_t$ contains rollouts, scores, reward events, deaths, selected rules, predicates and object positions; $D_t$ is the recent ESBM diff; $Q_t$ is QA history; $H_t$ is transition-prediction history; and $U_t$ is uncertain or contradictory mechanism memory.

Let $G(M)$ denote original-environment score. Given an adaptive QA benchmark $\mathcal{B}_t$ and a world-model probe suite $\mathcal{P}_t$, $A(M,\mathcal{B}_t)$ measures evidence-supported QA accuracy, whereas $W_{\mathrm{passive}}(M,\mathcal{P}_t)$ and $W_{\mathrm{active}}(M,\mathcal{P}_t)$ measure passive replay and active branch prediction accuracy. Score-only learning primarily maximizes $G(M)$; our goal is to learn ESBMs that improve score together with QA accuracy and executable world-model accuracy. Recovery under mechanism changes is then evaluated as a separate continuation setting, where the pre-change ESBM keeps optimizing after the changed mechanism is revealed through new interaction.

\begin{figure}[t]
\centering
\includegraphics[width=\textwidth]{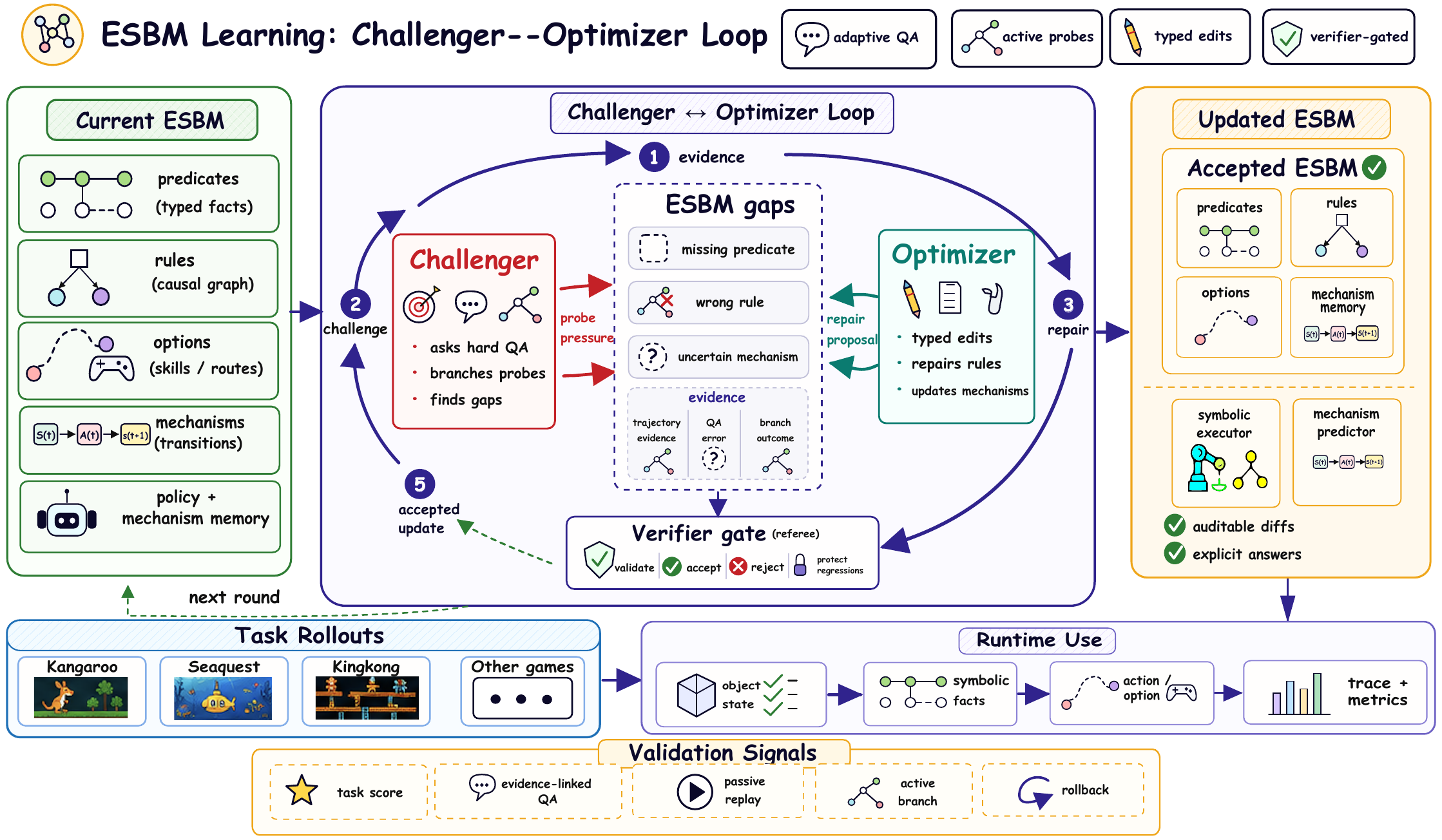}
\caption{\textbf{ESBM architecture with a Challenger--Optimizer training loop.}
The current ESBM executes as both a policy and an explicit mechanism model. Rollout evidence, QA mistakes and transition errors drive a challenger that applies adaptive questions and active world-model probes to expose missing predicates, incorrect rules and uncertain mechanisms. The optimizer responds with typed symbolic edits, which are filtered by a verifier gate before becoming accepted ESBM updates for the next training round.}
\label{fig:esbm-architecture}
\end{figure}

Figure~\ref{fig:esbm-architecture} summarizes the resulting architecture. The loop is adversarial in what it tests but constructive in what it accepts: the challenger turns rollout evidence and model errors into questions and active probes, the optimizer proposes typed local repairs to the ESBM, and a verifier gate accepts only updates that improve a validated dimension without violating protected regressions.

\section{Explicit Symbolic Behavioral Model}\label{sec:esbm}

An ESBM is the optimization object of the loop: an editable symbolic model that executes as a policy and also predicts abstract consequences of actions. The LLM is not the learned policy; it proposes local ESBM edits that are accepted only through validation. At runtime, $\Phi$ converts the raw observation and memory into symbolic facts, $\Pi$ selects a primitive action or option call, $O$ executes bounded multi-step procedures, and $K$ predicts how relevant symbolic state variables should change. Thus the policy is only one part of the model; the mechanism memory is the executable understanding layer that can be queried and repaired.

The same ESBM is used in three modes. For policy execution, the agent maps observation and memory to symbolic facts, selects a primitive action or bounded option, and updates memory after the option or action terminates. For question answering, the model exposes approved context from predicates, rules, options, mechanisms and selected trajectories, allowing an answerer to reason over explicit model content rather than hidden policy state. For world-model probing, $K$ receives a symbolic state and action sequence and predicts events, object changes, reward changes and terminal consequences. These modes share the same editable model, so a correction to a predicate, rule, option or mechanism can change behavior and understanding together.

Formally, at environment step $k$, the abstraction layer produces symbolic facts and the rule layer scores primitive actions and option calls:
\begin{align}
    z_k &= \Phi(o_k,m_k,K), \\
    s(u \mid z_k) &= \sum_{r \in \Pi} w_r\,
    \mathbf{1}\{\operatorname{body}(r) \subseteq z_k\}
    \mathbf{1}\{\operatorname{head}(r)=u\}, \\
    u_k &= \arg\max_{u \in \mathcal{A} \cup O} s(u \mid z_k).
\end{align}
Here $\mathcal{A}$ is the primitive action set, $O$ is the option library, and each rule contributes only when its body is satisfied by the current symbolic facts.

The state abstraction $\Phi$ is represented by typed predicates. Each predicate specifies typed variables, structured conditions and an output fact. Conditions include visibility, object attributes, spatial relations, distance thresholds, region membership and memory predicates. The rule policy $\Pi$ is a set of weighted clauses whose bodies are symbolic facts and whose heads are primitive actions or option calls. A rule stores its action, priority, weight, provenance and supporting or failing trajectories. This makes a behavioral change local: a rollout failure can be traced to a missing predicate, an over-broad condition, an incorrect action head or a bad weight.

Options allow the model to express behavior that should not be flattened into one-step rules. An option has typed inputs, preconditions, read/write memory fields, a termination condition, a maximum step bound and verifier checks. It may implement a short route, an interception routine or a bounded rescue sequence, but it cannot become unrestricted policy code because it must expose its contract and terminate within the declared bound.

Mechanism memory $K$ stores executable transition hypotheses. A mechanism contains trigger conditions, an action scope, predicted events, reward and life deltas, player-motion deltas, object deltas, confidence and trajectory evidence. Given symbolic state $z$ and an action or action sequence $\alpha$, the mechanism layer predicts
\begin{equation}
    K(z,\alpha) \rightarrow (\hat{e}, \Delta \hat{z}, \Delta \hat{r}, \Delta \hat{l}),
\end{equation}
where $\hat{e}$ denotes symbolic events, $\Delta \hat{z}$ object and relation changes, $\Delta \hat{r}$ reward change and $\Delta \hat{l}$ life or terminal-state change. This design makes understanding testable: the same mechanism entry can answer a QA item and can also be checked against a branched environment transition.

\section{Policy-adaptive QA and active world-model probes}\label{sec:generation}

The challenger is a probe generator conditioned on the current ESBM state. Its input is not only the latest trajectory, but also the current model version, recent symbolic diff, failure windows, reward events, deaths or stuck states, behavioral regressions, previous QA mistakes, uncertain mechanism entries and transition-prediction errors. Its output is a set of grounded QA items and active world-model probes that target concepts the current ESBM appears not to understand.

We write this generation step as
\begin{equation}
    (\mathcal{B}_t,\mathcal{P}_t) = C(S_t),
\end{equation}
where $C$ is the challenger, $\mathcal{B}_t$ is the adaptive QA set and $\mathcal{P}_t$ is the active world-model probe suite for the current ESBM state $S_t$.

The conditioning signals are chosen to make the benchmark policy-adaptive rather than static. Failure traces generate diagnosis questions about observed deaths, missed rewards or stuck behavior; symbolic diffs test whether a recent edit created a blind spot; previous QA mistakes focus new questions on concepts the model failed to support; transition errors select states where the mechanism layer predicts incorrectly; mechanism uncertainty probes missing or contradictory entries in $K$; and controlled environment modifications generate counterfactual and recovery probes. The detailed signal taxonomy is reported in the Supplementary Information.

QA items cover world structure, factual recognition, game mechanics, mechanistic reasoning, scoring and hazards, routes and strategy, and counterfactual reasoning. Each item stores a question, answer format, gold answer, target concept, model version and grounding evidence. Gold answers are accepted only when grounded in a trajectory event, a verified environment branch, an accepted mechanism entry or a declared environment modification. A separate answerer then receives the question and approved ESBM context: selected predicate and rule summaries, option contracts, mechanism entries, trajectory snippets and optional modification descriptions. It does not receive the challenger's hidden rationale or the gold answer. The answerer returns a structured answer with supporting evidence identifiers. QA scoring checks both answer correctness and evidence support, so an answer that is correct but unsupported by the ESBM context is marked as partial or failed.

For a question $b_i \in \mathcal{B}_t$, let $c_i(M)$ indicate answer correctness and $e_i(M)$ indicate whether the answer is supported by approved ESBM evidence. The benchmark score is
\begin{equation}
    A(M,\mathcal{B}_t) =
    \frac{1}{|\mathcal{B}_t|}
    \sum_{b_i \in \mathcal{B}_t} c_i(M)e_i(M).
\end{equation}

Active world-model probes test $K$ as an abstract simulator. From a checkpointed environment state, the system chooses a short action sequence and runs it in two branches. The real branch steps the environment and extracts observed symbolic events and final facts. The abstract branch applies $K$ to the same initial symbolic state and action sequence. The verifier compares predicted and observed events, reward changes, life changes, player motion and object changes. Action sequences come from three sources: observed failure windows, local counterfactuals that replace one or a few actions, and mechanism-targeted probes aimed at uncertain or contradictory entries in $K$.

For an active probe $p_i=(s_i,z_i,\alpha_i)$ with environment checkpoint $s_i$, symbolic state $z_i$ and action sequence $\alpha_i$, let $X$ extract symbolic outcomes from the real environment. Active world-model accuracy is
\begin{equation}
    W_{\mathrm{active}}(M,\mathcal{P}_t) =
    \frac{1}{|\mathcal{P}_t|}
    \sum_{p_i \in \mathcal{P}_t}
    \mathbf{1}\{d(K(z_i,\alpha_i), X(T(s_i,\alpha_i))) \leq \epsilon\},
\end{equation}
where $T$ is the environment transition function and $d$ compares predicted and observed symbolic outcomes.

Passive replay and active branching play different roles. Passive replay checks whether mechanism memory explains transitions already recorded in trajectories. Active branching checks whether the mechanism memory can predict consequences under selected interventions from a matched checkpoint. The second test is stricter because it prevents the model from merely rationalizing observed data after the fact.

\section{Joint ESBM optimization}\label{sec:learning}

Learning proceeds through repeated local edits to the ESBM. The optimizer may add or revise predicates, rule conditions, action heads, rule weights, bounded options or mechanism entries. Each proposed edit is evaluated as a candidate model rather than as a free-form explanation. The validation vector is
\begin{equation}
    V(M) = \big(G(M), A(M), W_{\mathrm{passive}}(M), W_{\mathrm{active}}(M)\big),
\end{equation}
where the terms denote task performance, QA accuracy, passive transition accuracy and active branch accuracy. No single scalar is treated as a complete proxy for success.

For a proposed update $M'_t$, we compare it with the current accepted model by
\begin{equation}
    \Delta V_t = V(M'_t)-V(M_t).
\end{equation}
With improvement thresholds $\tau$ and protected-regression tolerances $\rho$, the default acceptance condition is
\begin{equation}
    \exists j:\Delta V_{t,j}\geq \tau_j
    \quad\mathrm{and}\quad
    \forall k:\Delta V_{t,k}\geq -\rho_k .
\end{equation}
Mechanism repairs and nondominated held-out improvements are handled as explicit exceptions to this default rule.

The acceptance rule is thresholded rather than purely scalar. A candidate is acceptable if it improves at least one validated dimension while staying within regression tolerances on protected dimensions, repairs a verified mechanism error without making behavior nonviable, or becomes nondominated when evaluated on held-out probes. This permits a mechanism-only correction to enter the model even before it produces an immediate score gain, while still preventing ungrounded explanations from being accepted. The full optimization pseudocode is reported in Algorithm~\ref{alg:loop} in the Supplementary Information. Modified-environment recovery is not part of this per-iteration acceptance vector; it is evaluated by continuing the same Challenger--Optimizer loop after a controlled mechanism change and measuring how quickly the model adapts.

\section{Experiments}\label{sec:experiments}

The experiments are designed to answer four research questions:
\begin{itemize}
    \item \textbf{RQ1:} Can ESBM achieve competitive task performance rather than only producing interpretable explanations?
    \item \textbf{RQ2:} Does ESBM encode explicit game understanding that can be queried and supported by model evidence?
    \item \textbf{RQ3:} Does the mechanism layer form an executable world model that predicts held-out and counterfactual symbolic dynamics?
    \item \textbf{RQ4:} Does explicit game understanding improve recovery after controlled mechanism changes?
\end{itemize}
The main comparison uses three Atari-style JAXAtari games with distinct mechanisms: \texttt{kangaroo} for route planning and hazards, \texttt{seaquest} for oxygen and rescue mechanics, and \texttt{kingkong} for route adaptation and timing. Scores on additional games, including \texttt{pong}, are reported in the appendix as broader ESBM score context. The external reference methods are DQN and PPO as model-free score-optimizing baselines, NUDGE as a neurally guided differentiable logic policy learner, and BlendRL as a hybrid neural-symbolic policy learner \cite{dqn,ppo,nudge,blendrl}.

\begin{table}[t]
\caption{\textbf{Capability overview.} A check mark indicates that the capability is part of the method design; a circle indicates partial or indirect support. EvoSymbol$^*$ is the score-only self-evolving symbolic reference used in RQ1.}\label{tab:capability}
\scriptsize
\begin{tabular*}{\textwidth}{@{\extracolsep{\fill}}lcccccc@{}}
\toprule
Variant & Score & \shortstack{Explicit\\policy} & \shortstack{Agentic\\edits} & \shortstack{Adaptive\\QA} & \shortstack{Exec.\\WM} & \shortstack{Joint\\val.} \\
\midrule
DQN~\cite{dqn} & \cmark & \xmark & \xmark & \xmark & \xmark & \xmark \\
PPO~\cite{ppo} & \cmark & \xmark & \xmark & \xmark & \xmark & \xmark \\
NUDGE~\cite{nudge} & \cmark & \cmark & \xmark & \xmark & \xmark & \xmark \\
BlendRL~\cite{blendrl} & \cmark & \pmark & \xmark & \xmark & \pmark & \xmark \\
EvoSymbol$^*$ & \cmark & \cmark & \cmark & \xmark & \xmark & \xmark \\
ESBM & \cmark & \cmark & \cmark & \cmark & \cmark & \cmark \\
\botrule
\end{tabular*}
\end{table}

\subsection{Experimental setup}\label{subsec:setup}

Each main experiment corresponds to one research question and is reported on \texttt{kangaroo}, \texttt{seaquest} and \texttt{kingkong}. Additional game scores and protocol details are reported in the appendix. All ESBM variants use the same rollout budget, LLM budget and maximum model-update iterations within each game.

\subsection{Competitive task performance}

To answer RQ1, we compare the full ESBM with model-free, neuro-symbolic and score-only symbolic references. DQN, NUDGE and BlendRL provide external raw-score references where official or adapter-based results are available, whereas PPO, EvoSymbol and ESBM form the strict same-wrapper comparison. EvoSymbol isolates score-only symbolic self-evolution without QA or world-model validation. The main metric is mean game score over seeds; protocol details and external-reference settings are reported in the appendix.

\begin{table}[t]
\caption{\textbf{Task score.} Values are mean $\pm$ s.d. over evaluation seeds when available. Strict same-wrapper comparison is limited to PPO, EvoSymbol and ESBM; DQN, NUDGE and BlendRL provide external score references, with the \texttt{kingkong} BlendRL entry using a local semantic adapter because the official BlendRL benchmark reports DonkeyKong rather than this setting.}\label{tab:score}
\small
\renewcommand{\arraystretch}{1.08}
\setlength{\tabcolsep}{4pt}
\begin{tabular*}{\textwidth}{@{\extracolsep{\fill}}llrrr@{}}
\toprule
\multicolumn{2}{c}{Agent} & \multicolumn{3}{c}{Task score} \\
\cmidrule(lr){1-2}\cmidrule(lr){3-5}
Method & Type & \texttt{kangaroo} & \texttt{seaquest} & \texttt{kingkong} \\
\midrule
DQN & model-free & 2,696 & 2,794 & 1,638.5 $\pm$ 273.8 \\
PPO & model-free & 3,800 & 524 $\pm$ 8.94 & 1,250 $\pm$ 495 \\
\addlinespace[2pt]
NUDGE & logic-guided & 3,058 $\pm$ 25 & 64 $\pm$ 0 & 866 $\pm$ 56 \\
BlendRL & hybrid & 12,619 $\pm$ 132 & 4,204 $\pm$ 10 & 858.0 $\pm$ 379.2 \\
\addlinespace[2pt]
EvoSymbol$^*$ & score-only & 96,800 $\pm$ 0 & 8,442 $\pm$ 1,509 & 1,426 $\pm$ 503 \\
\rowcolor{hmmid}\textbf{ESBM} & \textbf{explicit} & \textbf{104,010 $\pm$ 32} & \textbf{11,716 $\pm$ 3,715} & \textbf{2,726 $\pm$ 977} \\
\botrule
\end{tabular*}
\par\smallskip\raggedright\scriptsize $^*$EvoSymbol is a score-only self-evolving symbolic agent without adaptive QA or executable world-model validation.\par
\end{table}

Within this mixed reference table, ESBM is the strongest reported entry on all three primary games. The strict same-wrapper comparison is between PPO, EvoSymbol and ESBM; EvoSymbol is also strong on \texttt{kangaroo}, but it separates score-only symbolic self-evolution from the QA- and world-model-validated ESBM objective.

\subsection{Explicit game understanding}

To answer RQ2, we freeze each accepted ESBM and evaluate it on a held-out game-understanding benchmark covering world structure, factual recognition, game mechanics, scoring and hazards, routes and strategy, and counterfactual reasoning. An answer is counted as evidence-supported only if the supporting predicate, rule, option, mechanism entry or trajectory evidence can be identified. Table~\ref{tab:understanding} therefore reports ESBM's evidence-linked answerability and audit fields, rather than treating methods without an explicit symbolic answer interface as failed QA respondents.

\begin{table}[t]
\caption{\textbf{Explicit understanding.} Values are policy-only answerability scores on evidence-linked QA. Cell shading highlights stronger evidence-supported understanding; \emph{Unknown} is reverse-coded because lower values indicate fewer unresolved answers.}\label{tab:understanding}
\footnotesize
\setlength{\tabcolsep}{5pt}
\begin{tabular*}{\textwidth}{@{\extracolsep{\fill}}lrrrrrr@{}}
\toprule
& \multicolumn{4}{c}{Evidence-supported answerability} & \multicolumn{2}{c}{Evidence audit} \\
\cmidrule(lr){2-5}\cmidrule(l){6-7}
Game & Overall & Factual & Mech. & CF & Coverage & Unknown \\
\midrule
\texttt{kangaroo} & \hmhighcell{0.949} & \hmhighcell{0.938} & \hmhighcell{0.949} & \hmhighcell{0.957} & \hmmidcell{0.739} & \hmhighcell{0.007} \\
\texttt{seaquest} & \hmlowcell{0.732} & \hmmidcell{0.808} & \hmlowcell{0.698} & \hmlowcell{0.786} & \hmlowcell{0.593} & \hmmidcell{0.046} \\
\texttt{kingkong} & \hmmidcell{0.855} & \hmhighcell{0.900} & \hmmidcell{0.832} & \hmhighcell{0.947} & \hmlowcell{0.671} & \hmwarncell{0.071} \\
\botrule
\end{tabular*}
\end{table}

The resulting QA scores show that the accepted ESBMs retain queryable evidence for most held-out understanding questions, with lower coverage on \texttt{seaquest} and \texttt{kingkong} marking where the symbolic model remains less complete.

\subsection{Executable world-model prediction}

To answer RQ3, we test whether ESBM mechanisms form an executable abstract world model. Starting from matched environment checkpoints, we execute the same action sequence in the real environment and in the ESBM mechanism layer. The prediction target includes symbolic next-state facts, object changes, collision and reward events, resource changes, life changes and terminal outcomes. The main metric is deliberately change-conditioned: static fields and unchanged objects are excluded so that trivial persistence does not appear strong by predicting no change. We compare ESBM with persistence, constant-dynamics and decision-tree world-model baselines under the same probe protocol.

\begin{table*}[t]
\caption{\textbf{World-model prediction.} Values are means $\pm$ s.d. under the online, data-budget-matched RQ3 protocol. Event F1 is micro-F1 over positive symbolic event labels. CF denotes restored counterfactual-branch change prediction under matched active probes; eligible target counts and denominator rules are reported in Table~\ref{tab:rq3denoms}.}\label{tab:worldmodel}
\footnotesize
\setlength{\tabcolsep}{4pt}
\begin{tabular*}{\textwidth}{@{\extracolsep{\fill}}llccccc@{}}
\toprule
& & \multicolumn{4}{c}{Observed-transition prediction} & \multicolumn{1}{c}{Active probe} \\
\cmidrule(lr){3-6}\cmidrule(l){7-7}
Game & Model & Fields & Event F1 & Rewards & Hazards & CF branch \\
\midrule
\multirow{4}{*}{\texttt{kangaroo}}
& Persistence & $0.00{\pm}0.00$ & $0.00{\pm}0.00$ & $0.00{\pm}0.00$ & $0.00{\pm}0.00$ & $0.00{\pm}0.00$ \\
& Const. dynamics & $0.18{\pm}0.04$ & $0.61{\pm}0.06$ & $0.10{\pm}0.05$ & $0.08{\pm}0.03$ & $0.06{\pm}0.03$ \\
& Decision-tree WM & $0.38{\pm}0.07$ & $0.74{\pm}0.05$ & $0.34{\pm}0.08$ & $0.29{\pm}0.07$ & $0.21{\pm}0.06$ \\
& \textbf{ESBM} & $\mathbf{0.72{\pm}0.05}$ & $\mathbf{0.84{\pm}0.04}$ & $\mathbf{0.71{\pm}0.06}$ & $\mathbf{0.68{\pm}0.06}$ & $\mathbf{0.61{\pm}0.07}$ \\
\addlinespace[2pt]
\multirow{4}{*}{\texttt{seaquest}}
& Persistence & $0.00{\pm}0.00$ & $0.00{\pm}0.00$ & $0.00{\pm}0.00$ & $0.00{\pm}0.00$ & $0.00{\pm}0.00$ \\
& Const. dynamics & $0.15{\pm}0.03$ & $0.58{\pm}0.07$ & $0.36{\pm}0.06$ & $0.06{\pm}0.03$ & $0.07{\pm}0.03$ \\
& Decision-tree WM & $0.29{\pm}0.06$ & $0.70{\pm}0.06$ & $0.42{\pm}0.08$ & $0.18{\pm}0.05$ & $0.17{\pm}0.05$ \\
& \textbf{ESBM} & $\mathbf{0.66{\pm}0.06}$ & $\mathbf{0.80{\pm}0.05}$ & $\mathbf{0.69{\pm}0.07}$ & $\mathbf{0.58{\pm}0.08}$ & $\mathbf{0.55{\pm}0.08}$ \\
\addlinespace[2pt]
\multirow{4}{*}{\texttt{kingkong}}
& Persistence & $0.00{\pm}0.00$ & $0.00{\pm}0.00$ & $0.00{\pm}0.00$ & $0.00{\pm}0.00$ & $0.00{\pm}0.00$ \\
& Const. dynamics & $0.13{\pm}0.04$ & $0.46{\pm}0.08$ & $0.08{\pm}0.04$ & $0.04{\pm}0.02$ & $0.05{\pm}0.03$ \\
& Decision-tree WM & $0.24{\pm}0.06$ & $0.63{\pm}0.07$ & $0.18{\pm}0.06$ & $0.12{\pm}0.05$ & $0.14{\pm}0.05$ \\
& \textbf{ESBM} & $\mathbf{0.58{\pm}0.08}$ & $\mathbf{0.74{\pm}0.06}$ & $\mathbf{0.52{\pm}0.09}$ & $\mathbf{0.43{\pm}0.09}$ & $\mathbf{0.38{\pm}0.10}$ \\
\botrule
\end{tabular*}
\end{table*}

Across games, ESBM outperforms persistence, constant-dynamics and decision-tree world-model baselines under the reported symbolic probe protocol on both observed transitions and active counterfactual branches. The lower absolute values on \texttt{kingkong} reflect the harder timing and moving-hazard dynamics rather than a missing probe protocol.

\subsection{Recovery after mechanism changes}

To answer RQ4, we test whether models recover after the modified mechanism is revealed through new interaction. The frozen modified-environment score $r_0$ is used only as the starting point for recovery, not as a separate main result. PPO continues gradient-based training in the modified environment, whereas ESBM continues the Challenger--Optimizer loop from the same pre-modification model. The main shifts are faster monkeys in \texttt{kangaroo} under the first-level/no-falling-coconut protocol, faster oxygen decay in \texttt{seaquest}, and faster bombs in first-level \texttt{kingkong}. Both methods receive the same post-modification interaction budget and are evaluated on the same seed set. For each method, $M_0$ denotes that method's pre-modification model. We report
\begin{equation}
    r_K =
    \frac{G_{\mathrm{mod}}(M_K)}{G_{\mathrm{orig}}(M_0)},
    \quad
    \Delta r = r_K-r_0 ,
\end{equation}
where $M_K$ is the recovered model after the matched post-modification budget.

\begin{table*}[t]
\caption{\textbf{Mechanism-change recovery.} Entries are score ratios normalized by original-environment score. Shaded cells highlight post-adaptation recovery and recovery gain after matched interaction.}\label{tab:recovery}
\footnotesize
\setlength{\tabcolsep}{5pt}
\begin{tabular*}{\textwidth}{@{\extracolsep{\fill}}llcccccc@{}}
\toprule
& & \multicolumn{3}{c}{PPO recovery} & \multicolumn{3}{c}{ESBM recovery} \\
\cmidrule(lr){3-5}\cmidrule(l){6-8}
Game & Mechanism shift & $r_0$ & $r_K$ & $\Delta r$ & $r_0$ & $r_K$ & $\Delta r$ \\
\midrule
\texttt{kangaroo} & faster monkeys & 0.54 & \hmlowcell{0.66} & \hmlowcell{+0.12} & 0.71 & \hmhighcell{0.94} & \hmhighcell{+0.23} \\
\texttt{seaquest} & faster oxygen decay & 0.46 & \hmlowcell{0.60} & \hmlowcell{+0.14} & 0.65 & \hmhighcell{0.90} & \hmhighcell{+0.25} \\
\texttt{kingkong} & faster bombs & 0.38 & \hmlowcell{0.53} & \hmlowcell{+0.15} & 0.61 & \hmmidcell{\textbf{0.86}} & \hmhighcell{+0.25} \\
\botrule
\end{tabular*}
\end{table*}

The recovery experiment therefore treats mechanism change as continued adaptation, not as another term in the per-iteration validation vector. ESBM recovers a larger fraction of the original score than PPO in all three controlled shifts.

\section{Related Work}\label{sec:related}

\subsection{Reflection-based LLM agents and program editing}

Reflection and self-feedback methods show that language agents can use task feedback or self-generated critique to improve behavior \cite{reflexion,selfrefine}. Executable skill repair and language-to-code control systems show that LLMs can write or revise executable policies from environment feedback \cite{voyager,codeaspolicies}. Reward-design methods use LLMs to generate reward programs for reinforcement learning \cite{eureka,text2reward}. These works motivate our optimizer, but they do not by themselves answer how to turn mechanism understanding into a policy-selection signal. In our setting, the LLM is a backend that proposes edits to an explicit model; the training signal comes from score, QA, and executable world-model validation.

\subsection{Self-evolving agent curricula}

Self-evolving agent frameworks reduce dependence on human-curated data by generating training tasks from the model's own frontier. Agent0 co-evolves a curriculum agent and an executor agent from the same base LLM, using tool-integrated reasoning so that a stronger executor pressures the curriculum agent to produce harder tool-aware tasks \cite{agent0}. This is related to our challenger--optimizer organization, but differs in both the optimization target and the validation object. Agent0 primarily improves a latent LLM agent through self-generated curricula. ESBM instead edits an explicit behavioral model in an interactive environment: the challenger generates evidence-grounded QA and active transition probes, and these probes become validation constraints over predicate, rule, option and mechanism edits.

\subsection{World-understanding evaluation}

Diagnostic QA and adversarial evaluation probe whether models can answer questions about objects, relations and reasoning cases beyond aggregate task success \cite{clevr,adversarialqa,adversarialvqa}. For interactive agents, QA-style probes are useful because task success can hide mechanistic misunderstandings, a failure mode related to shortcut learning in high-performing models \cite{shortcutlearning}. Our work builds on this view but moves QA from post-hoc evaluation into the learning loop. The generated questions are conditioned on the current model's failures, and QA accuracy is considered during ESBM selection.

\subsection{World models and transition prediction}

World-model and model-based reinforcement-learning approaches learn predictive models of environment dynamics for planning or policy improvement \cite{dyna,worldmodels,muzero,dreamerv2}. Our mechanism memory has a narrower but more inspectable role: it predicts symbolic events and state changes from symbolic state and action. The active validation protocol also differs from passive next-state prediction. It branches the real environment and the abstract ESBM world model from matched checkpoints under the same action sequence, then compares predicted and observed symbolic consequences. This makes mechanism memory an executable object that can be repaired, audited, and rejected when it contradicts observed dynamics.

\subsection{Dynamic and adversarial benchmarks}

Dynamic and adversarial benchmarks challenge models with examples designed against current failures \cite{dynabench,adversarialqa,anli,adversarialvqa}. Our benchmark is similarly adversarial, but it is conditioned on a changing ESBM rather than only on a frozen model. Generated questions and transition probes target rollout failures, recent symbolic edits, historical QA mistakes, and uncertain mechanism entries.

\subsection{Programmatic and symbolic policy learning}

Programmatic and symbolic reinforcement learning aims to produce interpretable policies \cite{pirl,nudge,blendrl}. Kintsugi further frames symbolic-agent improvement as verifier-gated knowledge-base induction: learned policy content is stored in YAML knowledge bases and updated through typed KBDiff operations applied by a proposer--applier--verifier loop \cite{kintsugi}. This is close to our use of typed local edits and verifier-gated acceptance. The difference is that ESBM treats the symbolic policy as only one part of a broader behavioral model. In our setting, predicates, rules and options are optimized together with evidence-grounded QA and active transition probes, so accepted edits can be connected not only to changed behavior but also to explicit answerability and executable mechanism consistency.

\section{Limitations}\label{sec:limitations}

The framework depends on the quality of generated probes. Poorly grounded or repetitive QA can create noise, and QA accuracy may not always correlate with game score. Active world-model validation also requires reliable checkpointing, symbolic state extraction, and action-sequence selection; a failed checkpoint replay can reduce usable probe coverage. Generator-answerer leakage must be controlled carefully. The symbolic vocabulary limits what the ESBM can express, and the option interface must remain bounded to avoid turning the model into unrestricted policy code. The current experiments isolate score-only symbolic self-evolution through EvoSymbol, but they are not a complete factorial ablation of every acceptance-rule component. Finally, the current setting uses Atari-style games as a controlled testbed; transferring the method to robotics or open-ended environments requires richer state abstraction and stronger verifiers.

\section{Conclusion}\label{sec:conclusion}

We introduced ESBM learning with adaptive questions and active world-model probes for learning explicit behavioral models in interactive environments. The method trains an ESBM that contains predicates, rules, options, and executable mechanism memory, rather than optimizing a standalone policy. A challenger generates grounded QA and active world-model probes from current failures, and an optimizer edits the ESBM under a multi-criterion acceptance rule that includes understanding checks. By making mechanism understanding executable and tied to policy updates, the framework separates high-score behavior from symbolically verifiable mechanism knowledge and provides a concrete mechanism for exposing brittle reward gains.

\backmatter

\bibliography{sn-bibliography}

\clearpage
\appendix
\setcounter{table}{0}
\setcounter{figure}{0}
\renewcommand{\thetable}{S\arabic{table}}
\renewcommand{\thefigure}{S\arabic{figure}}
\makeatletter
\@ifundefined{theHtable}{}{\renewcommand{\theHtable}{supp.\arabic{table}}}
\@ifundefined{theHfigure}{}{\renewcommand{\theHfigure}{supp.\arabic{figure}}}
\makeatother
\renewcommand{\tablename}{Supplementary Table}
\renewcommand{\figurename}{Supplementary Fig.}

\section{Supplementary Information}\label{app:supplementary}
\suppressfloats[t]

\subsection{ESBM schema summary}

The Explicit Symbolic Behavioral Model (ESBM) is the editable optimization object used throughout the study. It contains typed state abstractions, executable policy rules, bounded options and mechanism memory. The main text defines the model as
\begin{equation}
    M = (\Sigma, \Phi, \Pi, K, O),
\end{equation}
where $\Sigma$ is the symbolic vocabulary, $\Phi$ maps observations and memory into symbolic facts, $\Pi$ maps facts into primitive actions or option calls, $K$ stores executable mechanism knowledge, and $O$ stores bounded options. The appendix gives a component schema summary used to interpret the experimental artifacts.

\begin{table}[!h]
\caption{\textbf{ESBM schema.} Components define the symbolic vocabulary, abstraction, policy, options and mechanism memory used by the verifier during QA and world-model checks.}
\footnotesize
\setlength{\tabcolsep}{5pt}
\begin{tabular*}{\textwidth}{@{\extracolsep{\fill}}p{0.13\textwidth}p{0.34\textwidth}p{0.42\textwidth}@{}}
\toprule
Component & Contents & Validation role \\
\midrule
\textbf{$\Sigma$} & Entity types, object roles, region names, action names and symbolic fact names & Checks whether generated predicates and questions refer to valid concepts \\
\textbf{$\Phi$} & Typed predicates with structured conditions over observations, objects, memory and mechanism facts & Determines whether observations are converted into stable policy-relevant facts \\
\textbf{$\Pi$} & Weighted rules mapping facts to primitive actions or option calls & Determines action selection and links behavior changes to local symbolic diffs \\
\textbf{$O$} & Bounded option contracts with explicit inputs, termination conditions and allowed actions & Represents temporally extended behavior without unrestricted policy code \\
\textbf{$K$} & Mechanism entries, transition predictors, event predictors and natural-language rationales & Predicts symbolic consequences for QA and active world-model validation \\
\botrule
\end{tabular*}
\end{table}

\paragraph{Predicate entries.}
Each predicate entry specifies a name, typed variables, structured conditions and an output fact. Conditions include visibility, object attributes, spatial relations, distance thresholds, region membership, memory predicates and mechanism-derived facts. A predicate is accepted only if it type-checks, can be evaluated on stored trajectories and improves at least one validated dimension without violating protected regressions.

\paragraph{Rule entries.}
Each rule stores its action head, priority, weight, required facts, forbidden facts, provenance, supporting trajectories and failing trajectories. At runtime, all satisfied rules contribute scores to primitive actions or option calls. Rule edits include adding a rule, deleting a rule, changing a condition, changing an action head, and adjusting a priority or weight.

\paragraph{Option entries.}
Options represent bounded procedures that may require short-horizon memory, repeated actions or local search. An option must declare its inputs, allowed primitive actions, termination condition, timeout and failure condition. Options are not unrestricted task solutions: they are executable subroutines whose preconditions and outcomes remain visible to the ESBM validator.

\paragraph{Mechanism entries.}
Mechanism memory contains executable predictions and natural-language rationales. The executable part maps a symbolic state and an action or action sequence to predicted events, object changes, reward changes and life changes. The rationale part explains why the transition is expected and which trajectories support it. A mechanism entry may improve QA accuracy before it improves score, but it must pass passive replay or active branch validation before it is treated as reliable.

\subsection{Adaptive QA benchmark design}

The adaptive QA benchmark is generated from the current ESBM state rather than from a fixed question list. Its purpose is to expose model-specific blind spots and to produce reusable tests of environment understanding. Questions are generated from rollout failures, high-value events, recent symbolic edits, previous QA mistakes, uncertain mechanism entries and active world-model discrepancies.

\begin{table}[t]
\caption{\textbf{Probe signals.} Each signal selects a different source of pressure for QA generation or active world-model validation in each update round.}
\footnotesize
\setlength{\tabcolsep}{5pt}
\begin{tabular*}{\textwidth}{@{\extracolsep{\fill}}p{0.25\textwidth}p{0.66\textwidth}@{}}
\toprule
Signal & Role in probe generation \\
\midrule
\textbf{Failure trace} & Generates diagnosis questions about observed deaths, missed rewards, stalled states and repeated wrong actions \\
\textbf{Symbolic diff} & Tests whether a recent predicate, rule, option or mechanism edit introduced a new blind spot \\
\textbf{QA mistakes} & Focuses new questions on concepts that the current model failed to answer with evidence \\
\textbf{Transition errors} & Selects states and action windows where the executable world model predicts incorrectly \\
\textbf{Mechanism uncertainty} & Probes missing, low-confidence or contradictory mechanism entries in $K$ \\
\textbf{Environment modification} & Generates counterfactual and recovery probes under declared mechanism changes \\
\botrule
\end{tabular*}
\end{table}

\begin{table}[t]
\caption{\textbf{QA categories.} Categories define the target ability and the evidence used to ground generated answers during held-out evaluation.}
\footnotesize
\setlength{\tabcolsep}{5pt}
\begin{tabular*}{\textwidth}{@{\extracolsep{\fill}}p{0.21\textwidth}p{0.31\textwidth}p{0.37\textwidth}@{}}
\toprule
Category & Target ability & Typical grounding evidence \\
\midrule
\textbf{World structure} & Entities, roles, regions and persistent spatial layout & Object lists, region definitions and stable layout observations \\
\textbf{Factual recognition} & Entity descriptions, object roles and visible state facts & Trajectory frames and symbolic state snapshots \\
\textbf{Game mechanics} & Interaction rules, transition dynamics and resource behavior & State-action-next-state transitions \\
\textbf{Scoring and hazards} & Reward sources, failure causes and danger-aware items & Reward events, death events and hazard contacts \\
\textbf{Routes and strategy} & Subgoals, high-value paths and route abstractions & Successful and failed trajectory segments \\
\textbf{Counterfactual reasoning} & Consequences under modified assumptions or action changes & Verified branches and declared environment modifications \\
\botrule
\end{tabular*}
\end{table}

\paragraph{Question generation.}
For each iteration, the challenger receives the current model summary, recent rollout traces, accepted and rejected diffs, QA history and mechanism-validation history. It proposes questions with a target concept, answer format, gold answer, difficulty estimate and grounding evidence. A generated question is retained only if its gold answer is grounded in a trajectory event, a verified active branch, an accepted mechanism entry or a declared environment modification.

\paragraph{Question answering.}
The answerer receives the question and approved ESBM context: selected predicate summaries, rule summaries, option contracts, mechanism entries, trajectory snippets and optional environment-modification descriptions. It does not receive the challenger's hidden rationale or the gold answer. The answerer returns a structured answer and evidence identifiers. QA scoring checks answer correctness and whether the answer is supported by the provided ESBM context.

\subsection{Active world-model probes}

Active world-model probes test whether mechanism memory can predict symbolic consequences under interventions. A probe begins from a matched checkpoint $(s_i, z_i)$, where $s_i$ is the environment state and $z_i$ is the ESBM symbolic state. The same action sequence is applied to the real environment and to the abstract ESBM world model:
\begin{equation}
    s_{i+k} = T(s_i, a_{i:i+k}), \qquad
    \hat{z}_{i+k} = K(z_i, a_{i:i+k}).
\end{equation}
The validator compares observed and predicted symbolic events, object positions, reward changes, life changes and resource changes.

\begin{table}[t]
\caption{\textbf{Active branches.} Sequence families define how action branches are constructed and what causal property they test in the ESBM world model.}
\footnotesize
\setlength{\tabcolsep}{5pt}
\begin{tabular*}{\textwidth}{@{\extracolsep{\fill}}p{0.22\textwidth}p{0.34\textwidth}p{0.34\textwidth}@{}}
\toprule
Sequence family & Construction & Purpose \\
\midrule
\textbf{Observed replay} & Reuse an action segment from a real trajectory & Tests whether $K$ can reproduce known dynamics \\
\textbf{Local counterfactual} & Change one or a few actions near a failure or reward event & Tests causal sensitivity around important events \\
\textbf{Policy branch} & Follow the current ESBM policy from a stored checkpoint & Tests whether model predictions match its own behavior distribution \\
\textbf{Mixed branch} & Combine observed prefixes with counterfactual suffixes & Tests generalization beyond the exact recorded rollout \\
\botrule
\end{tabular*}
\end{table}

\subsection{RQ3 metric construction and baselines}

The RQ3 world-model results in Table~\ref{tab:worldmodel} are computed from the final accepted ESBM artifacts for the three primary games. For each game and seed, we first collect evaluation rollouts under the same 20,000-frame cap used for score reporting. The validator then builds two probe sets. Passive probes are held-out state-action windows from recorded trajectories. Active probes restore a matched environment checkpoint and execute an observed, counterfactual, policy or mixed action sequence in both the real environment and the ESBM mechanism layer. Scores are computed per seed and reported as mean $\pm$ s.d. across seeds.

\begin{table}[t]
\caption{\textbf{World-model metrics.} Each metric is scored only on eligible probe targets, so unchanged fields and unavailable branch replays do not create trivial successes.}
\footnotesize
\setlength{\tabcolsep}{5pt}
\begin{tabular*}{\textwidth}{@{\extracolsep{\fill}}p{0.18\textwidth}p{0.38\textwidth}p{0.34\textwidth}@{}}
\toprule
Metric & Eligible target & Correct prediction \\
\midrule
\textbf{Fields} & Symbolic scalar or relational fields that change within the probe window & Predicted final value matches the extracted final symbolic value \\
\textbf{Event F1} & Positive non-bookkeeping event labels emitted by the extractor and the mechanism layer & Micro-F1 between predicted and observed positive event labels \\
\textbf{Rewards} & Windows with non-zero reward change or declared reward-relevant mechanism triggers & Predicted reward delta matches the observed reward delta within the declared tolerance \\
\textbf{Hazards} & Windows involving hazard contact, avoidance, life loss or terminal-risk mechanisms & Predicted hazard, life or terminal consequence matches the observed outcome \\
\textbf{CF branch} & Restored active branches with a valid checkpoint and a non-empty changed target set & Aggregate changed-target prediction over the restored branch is correct \\
\botrule
\end{tabular*}
\end{table}

The event-F1 column is evaluated from explicit symbolic event predictions rather than inferred from the reward or hazard columns. Typical event labels include reward collection, hazard contact, rescue or resource events, object removal, projectile interactions and terminal or life-loss events, depending on the game. True no-event negatives do not contribute to this micro-F1 score. If a probe does not emit an eligible target for a metric, that probe is excluded from that metric's denominator rather than being counted as either a success or a failure.

\begin{table*}[t]
\caption{\textbf{RQ3 denominators.} Counts are totals across evaluation seeds after excluding static fields, empty event windows and failed checkpoint restores; parentheses report the mean eligible count per seed.}\label{tab:rq3denoms}
\scriptsize
\setlength{\tabcolsep}{3pt}
\begin{tabular*}{\textwidth}{@{\extracolsep{\fill}}lccccccc@{}}
\toprule
Game & Seeds & Passive & Active & Fields & Event labels & Rewards & Hazards \\
\midrule
\texttt{kangaroo} & 5 & 1,280 (256) & 150 (30) & 3,920 (784) & 562 (112) & 457 (91) & 543 (109) \\
\texttt{seaquest} & 5 & 1,280 (256) & 148 (30) & 5,940 (1,188) & 842 (168) & 711 (142) & 538 (108) \\
\texttt{kingkong} & 5 & 1,280 (256) & 142 (28) & 6,210 (1,242) & 1,036 (207) & 486 (97) & 672 (134) \\
\botrule
\end{tabular*}
\end{table*}

Table~\ref{tab:rq3denoms} reports the denominators used for the mean scores in Table~\ref{tab:worldmodel}. Field targets are individual changed symbolic fields, event-label targets are positive non-bookkeeping event labels, reward targets are non-zero reward or resource changes, and hazard targets are changed hazard, life-loss or terminal-risk facts. Active-branch denominators count restored branches with at least one changed target; failed restores and empty branches are excluded before any model is scored. All baselines and ESBM are evaluated on exactly the same eligible targets for a given game and seed.

All RQ3 baselines use the same symbolic state, action and probe budget available to the ESBM validator. The persistence baseline copies the initial symbolic state forward and predicts no events, no reward change and no hazard change; because static fields are excluded, it is a check against trivial no-change prediction. The constant-dynamics baseline predicts the most frequent observed transition, event, reward and hazard outcome for each game and action class from the training windows. The decision-tree world model is trained on the same symbolic state-action windows, using symbolic facts and action identifiers as features, and is evaluated on the held-out passive windows and restored active branches without access to their outcomes. This keeps the comparison data-budget matched while separating executable mechanism memory from generic symbolic transition fitting.

\subsection{ESBM optimization and version control}

Each accepted ESBM version is stored as a complete policy artifact plus a diff from the previous accepted version. Candidate updates may modify predicates, rules, options and mechanisms. The optimizer may use rollout failures, QA failures and transition-prediction failures as evidence. The candidate is accepted if it improves at least one validated dimension, remains within regression tolerances on protected dimensions, repairs a verified mechanism error without making behavior nonviable, or becomes nondominated on held-out probes.

\begin{algorithm}[h]
\caption{\textbf{Optimization loop.} Candidate updates are tested against score, QA and executable world-model signals before becoming the next accepted model for the following round.}\label{alg:loop}
\begin{algorithmic}[1]
\Require Initial ESBM $M_0$, rollout budget, QA budget, active-probe budget
\For{$t = 0, 1, \ldots, T-1$}
    \State Run rollout with current model $M_t$
    \State Build $S_t$ from score, traces, diffs, QA mistakes, and transition errors
    \State Generate QA benchmark $\mathcal{B}_t$ and probe suite $\mathcal{P}_t$
    \State Evaluate $A(M_t,\mathcal{B}_t)$, $W_{\mathrm{passive}}(M_t,\mathcal{P}_t)$, and $W_{\mathrm{active}}(M_t,\mathcal{P}_t)$
    \State Propose a local ESBM update $M'_t$
    \State Evaluate $G(M'_t)$, $A(M'_t,\mathcal{B}_t)$, $W_{\mathrm{passive}}(M'_t,\mathcal{P}_t)$, and $W_{\mathrm{active}}(M'_t,\mathcal{P}_t)$
    \If{$M'_t$ improves at least one validation dimension and no protected dimension regresses beyond tolerance}
        \State Accept $M_{t+1} \gets M'_t$
    \ElsIf{$M'_t$ repairs a verified mechanism error and remains behaviorally viable}
        \State Accept $M_{t+1} \gets M'_t$
    \ElsIf{$M'_t$ is nondominated on the validation vector and improves held-out probes}
        \State Accept $M_{t+1} \gets M'_t$
    \Else
        \State Reject update and keep $M_{t+1} \gets M_t$
    \EndIf
\EndFor
\end{algorithmic}
\end{algorithm}

\begin{table}[t]
\caption{\textbf{Typed edits.} Each edit type has execution, grounding or regression checks that prevent unsupported symbolic changes from entering the accepted model.}
\footnotesize
\setlength{\tabcolsep}{5pt}
\begin{tabular*}{\textwidth}{@{\extracolsep{\fill}}p{0.20\textwidth}p{0.35\textwidth}p{0.35\textwidth}@{}}
\toprule
Edit type & Examples & Required checks \\
\midrule
\textbf{Predicate edit} & Add, remove or change a typed predicate condition & Type check, replay evaluation and regression check \\
\textbf{Rule edit} & Add, remove or reweight a rule; change an action head & Runtime execution, score validation and protected behavior check \\
\textbf{Option edit} & Add or change a bounded option contract & Timeout, termination, action-set and rollback checks \\
\textbf{Mechanism edit} & Add or revise event or transition predictions & Passive replay or active branch validation \\
\textbf{QA edit} & Add generated QA item or revise held-out split & Gold grounding, leakage check and coverage check \\
\botrule
\end{tabular*}
\end{table}

\subsection{Experimental protocol}

All ESBM variants use the same rollout budget, LLM budget, maximum number of update iterations and random seeds within each environment. The main-text comparison uses three primary Atari-style JAXAtari games selected to test route planning and hazards, oxygen and rescue mechanics, and route adaptation under timing constraints. Additional environments are reported in the appendix as ESBM score results, not as part of the main score comparison.

\begin{table*}[t]
\caption{\textbf{Evaluation axes.} Main-text games are evaluated on score and understanding metrics; appendix-only games are reported as ESBM score context beyond the primary comparison.}
\footnotesize
\setlength{\tabcolsep}{3pt}
\begin{tabular*}{\textwidth}{@{\extracolsep{\fill}}p{0.10\textwidth}p{0.15\textwidth}p{0.31\textwidth}p{0.36\textwidth}@{}}
\toprule
Scope & Game & Primary mechanism & Reported metrics \\
\midrule
\hmhighcell{Main} & \texttt{kangaroo} & Route planning, ladders, hazards and scoring objects & Score, route QA, hazard QA, active branch accuracy and modified-environment score \\
\hmhighcell{Main} & \texttt{seaquest} & Oxygen, rescue, enemies and shooting dynamics & Score, oxygen QA, rescue QA, transition accuracy and modified-environment score \\
\hmhighcell{Main} & \texttt{kingkong} & Route adaptation, timing and moving hazards & Score, route QA, mechanism QA and modified-environment score \\
\hmmidcell{Appendix} & Broader JAXAtari set & Varied Atari-style mechanisms & Reported ESBM scores across additional games \\
\botrule
\end{tabular*}
\end{table*}

\subsection{Supplementary numerical results}

This section reports the score and QA artifacts used in the main-text tables, together with ESBM scores for additional JAXAtari environments. Published DQN, NUDGE and BlendRL values in Table~\ref{tab:score} follow the raw-score evaluation protocol reported in the BlendRL appendix where the same Atari game is available \cite{blendrl}. The official BlendRL Atari comparison reports \texttt{kangaroo}, \texttt{seaquest} and DonkeyKong; because our third primary game is the JAXAtari \texttt{kingkong} setting rather than DonkeyKong, the \texttt{kingkong} DQN, NUDGE and BlendRL entries use local first-level JAXAtari adapter runs. ESBM scores are evaluated under a fixed 20,000 raw-frame cap to prevent unbounded score accumulation in non-terminating policies, which is particularly important for \texttt{kangaroo}. EvoSymbol denotes score-only self-evolving symbolic-agent runs from the same score-record artifacts. Strict same-wrapper comparisons therefore remain limited to the in-protocol PPO, EvoSymbol and ESBM runs, whereas external DQN, NUDGE and BlendRL entries provide score context. The additional-game scores below are appendix-only score context; they are not used to support the RQ2--RQ4 mechanism-understanding claims and do not include the QA or world-model evaluation artifacts reported for the three primary games.

\begin{table*}[t]
\caption{\textbf{Additional scores.} Scores are appendix-only ESBM score context at 20,000 raw frames; they are not used for the QA, world-model or recovery claims. The three primary games are omitted to avoid duplicating the main result.}
\scriptsize
\setlength{\tabcolsep}{3pt}
\begin{tabular*}{\textwidth}{@{\extracolsep{\fill}}p{0.25\textwidth}r p{0.25\textwidth}r@{}}
\toprule
Game & ESBM score & Game & ESBM score \\
\midrule
\texttt{alien} & 290 & \texttt{lasergates} & 1,830 \\
\texttt{asterix} & $96,600 \pm 11,036$ & \texttt{namethisgame} & $5,070 \pm 1,096$ \\
\texttt{asteroids} & $4,774 \pm 2,946$ & \texttt{phoenix} & 3,080 \\
\texttt{atlantis} & $56,820 \pm 19,463$ & \texttt{pong} & 31.0 \\
\texttt{bankheist} & 4,185,308 & \texttt{riverraid} & 9,630 \\
\texttt{bankheist\_no\_\allowbreak{}portals} & $70,632 \pm 5.48$ & \texttt{sir\_lancelot} & $5,302 \pm 8,560$ \\
\texttt{berzerk} & $4,400 \pm 377$ & \texttt{skiing} & 20.0 \\
\texttt{blackjack} & $204 \pm 4.53$ & \texttt{slotmachine} & $270 \pm 450$ \\
\texttt{breakout} & 206 & \texttt{spaceinvaders} & $1,097 \pm 670$ \\
\texttt{centipede} & $7,401 \pm 3,996$ & \texttt{spacewar} & 4.00 \\
\texttt{choppercommand} & $22,540 \pm 462$ & \texttt{surround} & 10.0 \\
\texttt{enduro} & 105 & \texttt{tennis} & 0.00 \\
\texttt{fishingderby} & $103 \pm 1$ & \texttt{tetris} & $113 \pm 23.4$ \\
\texttt{flagcapture} & $16.6 \pm 1.52$ & \texttt{timepilot} & $12,460 \pm 3,210$ \\
\texttt{freeway} & 80.0 & \texttt{tron} & $494 \pm 269$ \\
\texttt{frostbite} & $3,426 \pm 972$ & \texttt{turmoil} & $12,022 \pm 643$ \\
\texttt{galaxian} & 4,260 & \texttt{videocheckers} & $6.40 \pm 3.44$ \\
\texttt{hangman} & $168 \pm 3.36$ & \texttt{videocube} & 953 \\
\texttt{haunted\_house} & $113,709 \pm 34,403$ & \texttt{videopinball} & $1,842 \pm 1,307$ \\
\texttt{human\_cannonball} & 7.00 & \texttt{wordzapper} & $105,058 \pm 163$ \\
\texttt{klax} & $2,459 \pm 2,964$ & & \\
\botrule
\end{tabular*}
\end{table*}

\begin{table*}[t]
\caption{\textbf{Primary artifacts.} Kangaroo uses the first-level/no-falling-coconut protocol; Kingkong uses the first-level-only protocol for score and QA reporting.}
\footnotesize
\setlength{\tabcolsep}{5pt}
\begin{tabular*}{\textwidth}{@{\extracolsep{\fill}}llccccc@{}}
\toprule
& & \multicolumn{2}{c}{Evaluation budget} & \multicolumn{3}{c}{Artifact outcome} \\
\cmidrule(lr){3-4}\cmidrule(l){5-7}
Game & Protocol/source & Seeds & Steps & Score & QA & Coverage \\
\midrule
\texttt{kangaroo} & \texttt{iter\_013} accepted & 5 & 20,000 & \textbf{104,010 $\pm$ 32} & \hmhighcell{0.949} & \hmmidcell{0.739} \\
\texttt{seaquest} & \texttt{iter\_026} accepted & 5 & 20,000 & \textbf{11,716 $\pm$ 3,715} & \hmlowcell{0.732} & \hmlowcell{0.593} \\
\texttt{kingkong} & \texttt{iter\_027} accepted & 5 & 20,000 & \textbf{2,726 $\pm$ 977} & \hmmidcell{0.855} & \hmlowcell{0.671} \\
\botrule
\end{tabular*}
\end{table*}

\begin{table*}[t]
\caption{\textbf{QA accumulation.} The same final QA benchmark is answered by earlier and later accepted policies to show how answerability grows over the accepted ESBM sequence.}
\footnotesize
\setlength{\tabcolsep}{5pt}
\begin{tabular*}{\textwidth}{@{\extracolsep{\fill}}lcrrrrr@{}}
\toprule
& \multicolumn{1}{c}{Benchmark} & \multicolumn{4}{c}{Answerability by accepted version} & \multicolumn{1}{c}{Net gain} \\
\cmidrule(lr){2-2}\cmidrule(lr){3-6}\cmidrule(l){7-7}
Game & QA & Start & Mid & Late & Final & Gain \\
\midrule
\texttt{kangaroo} & 138 & 0.000 & 0.138 & 0.471 & \hmhighcell{0.949} & \hmhighcell{+0.949} \\
\texttt{seaquest} & 194 & 0.000 & 0.180 & 0.701 & \hmlowcell{0.732} & \hmlowcell{+0.732} \\
\texttt{kingkong} & 310 & 0.000 & 0.587 & 0.748 & \hmmidcell{0.855} & \hmmidcell{+0.855} \\
\botrule
\end{tabular*}
\end{table*}

\begin{table}[t]
\caption{\textbf{World-model summary.} Values match the ESBM rows in Table~\ref{tab:worldmodel} for direct cross-reference with the main-text comparison.}
\footnotesize
\setlength{\tabcolsep}{5pt}
\begin{tabular*}{\textwidth}{@{\extracolsep{\fill}}lccccc@{}}
\toprule
Game & Fields & Event F1 & Rewards & Hazards & CF branch \\
\midrule
\texttt{kangaroo} & $0.72{\pm}0.05$ & $0.84{\pm}0.04$ & $0.71{\pm}0.06$ & $0.68{\pm}0.06$ & $0.61{\pm}0.07$ \\
\texttt{seaquest} & $0.66{\pm}0.06$ & $0.80{\pm}0.05$ & $0.69{\pm}0.07$ & $0.58{\pm}0.08$ & $0.55{\pm}0.08$ \\
\texttt{kingkong} & $0.58{\pm}0.08$ & $0.74{\pm}0.06$ & $0.52{\pm}0.09$ & $0.43{\pm}0.09$ & $0.38{\pm}0.10$ \\
\botrule
\end{tabular*}
\end{table}

\subsection{Recovery under modified environments}

Modified environments test whether an ESBM can revise a reusable behavioral model after a mechanism change rather than only relying on the original score policy. Each modification changes a specific mechanism while preserving the task identity. The table reports the initial modified-environment score and recovery after a matched post-modification interaction budget. PPO is evaluated by the same recovery protocol, but without QA or explicit world-model metrics. Table~\ref{tab:recoveryprotocol} summarizes the seed counts and frame caps used for the reported ratios.

\begin{table}[t]
\caption{\textbf{Recovery protocol.} Each method is adapted under the same post-modification frame budget within a game and then evaluated with the same raw-frame cap used for score reporting.}\label{tab:recoveryprotocol}
\footnotesize
\setlength{\tabcolsep}{4pt}
\begin{tabular*}{\textwidth}{@{\extracolsep{\fill}}lccc@{}}
\toprule
Game & Seeds & Post-mod budget & Evaluation cap \\
\midrule
\texttt{kangaroo} & 5 & 20,000 frames/seed & 20,000 frames/seed \\
\texttt{seaquest} & 5 & 20,000 frames/seed & 20,000 frames/seed \\
\texttt{kingkong} & 5 & 20,000 frames/seed & 20,000 frames/seed \\
\botrule
\end{tabular*}
\end{table}

\begin{table*}[t]
\caption{\textbf{Recovery details.} Score ratios are normalized by original-environment performance. Shaded cells mark the recovered score ratio and recovery gain after adaptation.}
\footnotesize
\setlength{\tabcolsep}{5pt}
\begin{tabular*}{\textwidth}{@{\extracolsep{\fill}}p{0.16\textwidth}p{0.26\textwidth}p{0.12\textwidth}cccc@{}}
\toprule
& & & \multicolumn{4}{c}{Score ratio} \\
\cmidrule(l){4-7}
Game & Shift & Method & Orig. & Init. & Recov. & Gain \\
\midrule
\texttt{kangaroo} & faster monkeys & PPO & 1.00 & 0.54 & \hmlowcell{0.66} & \hmlowcell{+0.12} \\
\texttt{kangaroo} & faster monkeys & \textbf{ESBM} & 1.00 & 0.71 & \hmhighcell{0.94} & \hmhighcell{+0.23} \\
\texttt{seaquest} & faster oxygen decay & PPO & 1.00 & 0.46 & \hmlowcell{0.60} & \hmlowcell{+0.14} \\
\texttt{seaquest} & faster oxygen decay & \textbf{ESBM} & 1.00 & 0.65 & \hmhighcell{0.90} & \hmhighcell{+0.25} \\
\texttt{kingkong} & faster bombs & PPO & 1.00 & 0.38 & \hmlowcell{0.53} & \hmlowcell{+0.15} \\
\texttt{kingkong} & faster bombs & \textbf{ESBM} & 1.00 & 0.61 & \hmmidcell{\textbf{0.86}} & \hmhighcell{+0.25} \\
\botrule
\end{tabular*}
\end{table*}

\subsection{Additional limitations}

The ESBM framework depends on reliable symbolic extraction, checkpoint replay and probe grounding. If the extractor misses important objects or if active branches cannot be replayed reliably, mechanism validation can be incomplete. QA generation can also introduce bias if questions repeatedly target easy concepts or leak the gold answer to the answerer. These risks are monitored through held-out QA, grounding checks, generator-answerer separation and probe-diversity statistics.

\end{document}